%% file: main.tex
\documentclass[sigconf]{acmart}

\usepackage{amsmath}
\usepackage{bm}
\usepackage{algorithm}
\usepackage{algpseudocode}
\usepackage{amsmath}
\usepackage{multirow}
\usepackage{tikz}
\usepackage{enumitem}

\usepackage{indentfirst}
\usepackage{subfigure}
\usepackage{epsfig}
\usepackage{epstopdf}
\usepackage{graphicx}
\usepackage{url}
\usepackage{xspace}
\usepackage{booktabs}
\usepackage{subeqnarray}
\usepackage{subcaption}
\usepackage{color}

\newcommand{\paratitle}[1]{\vspace{1.5ex}\noindent\textbf{#1}}

\newcommand{\ignore}[1]{}

\AtBeginDocument{%
  }

\copyrightyear{2025}
\acmYear{2025}
\setcopyright{acmlicensed}\acmConference[SIGIR '25]{Proceedings of the 48th International ACM SIGIR Conference on Research and Development in Information Retrieval}{July 13--18, 2025}{Padua, Italy}
\acmBooktitle{Proceedings of the 48th International ACM SIGIR Conference on Research and Development in Information Retrieval (SIGIR '25), July 13--18, 2025, Padua, Italy}
\acmDOI{10.1145/3726302.3730112}
\acmISBN{979-8-4007-1592-1/2025/07}



\begin{document}

% \title{Elicitation and Contestation: Interpreting LLM-based RAG from Knowledge Streaming Perspective}
\title{Unveiling Knowledge Utilization Mechanisms in LLM-based Retrieval-Augmented Generation}


\author{Yuhao Wang} \authornote{Equal Contributions.}
\authornote{The work was done during the internship at Baidu.}
\email{yh.wang500@outlook.com}
\affiliation{%
  \institution{GSAI, Renmin University of China}
  \city{Beijing}
  \country{China}}

\author{Ruiyang Ren} \authornotemark[1]
\email{reyon_ren@outlook.com}
\affiliation{%
  \institution{GSAI, Renmin University of China}
  \city{Beijing}
  \country{China}}

\author{Yucheng Wang}
\email{wangyucheng01@baidu.com}
\affiliation{%
  \institution{Baidu Inc.}
  \city{Beijing}
  \country{China}}

\author{Wayne Xin Zhao}\authornote{Corresponding Authors.}
\email{batmanfly@gmail.com}
\affiliation{%
  \institution{GSAI, Renmin University of China}
  \city{Beijing}
  \country{China}}

\author{Jing Liu} \authornotemark[3]
\email{liujing46@baidu.com}
\affiliation{%
  \institution{Baidu Inc.}
  \city{Beijing}
  \country{China}}
  
\author{Hua Wu}
\email{wu_hua@baidu.com}
\affiliation{%
  \institution{Baidu Inc.}
  \city{Beijing}
  \country{China}}

\author{Haifeng Wang}
\email{wanghaifeng@baidu.com}
\affiliation{%
  \institution{Baidu Inc.}
  \city{Beijing}
  \country{China}}

% \affiliation{%
%   \institution{Baidu Inc.}
%   \city{Beijing}


\renewcommand{\shortauthors}{Yuhao Wang et al.}

\begin{abstract}
Considering the inherent limitations of parametric knowledge in large language models (LLMs), retrieval-augmented generation (RAG) is widely employed to expand their knowledge scope.
Since RAG has shown promise in knowledge-intensive tasks like open-domain question answering, its broader application to complex tasks and intelligent assistants has further advanced its utility. Despite this progress, the underlying knowledge utilization mechanisms of LLM-based RAG remain underexplored. In this paper, we present a systematic investigation of the intrinsic mechanisms by which LLMs integrate internal (parametric) and external (retrieved) knowledge in RAG scenarios. 
Specially, we employ knowledge stream analysis at the macroscopic level, and investigate the function of individual modules at the microscopic level.
Drawing on knowledge streaming analyses, we decompose the knowledge utilization process into four distinct stages within LLM layers: knowledge refinement, knowledge elicitation, knowledge expression, and knowledge contestation. We further demonstrate that the relevance of passages guides the streaming of knowledge through these stages. 
At the module level, we introduce a new method, {k}nowledge {a}ctivation probability {e}ntropy~({KAPE}) for neuron identification associated with either internal or external knowledge.
By selectively deactivating these neurons, we achieve targeted shifts in the LLM’s reliance on one knowledge source over the other.
Moreover, we discern complementary roles for multi-head attention and multi-layer perceptron layers during knowledge formation.
These insights offer a foundation for improving interpretability and reliability in retrieval-augmented LLMs, paving the way for more robust and transparent generative solutions in knowledge-intensive domains.
\end{abstract}

\maketitle

\input{sec/sec-intro}

\input{sec/sec-preliminaries}
\input{sec/sec-analysis}
\input{sec/sec-indepth}

\input{sec/sec-conclusion}

\begin{acks}
This work was partially supported by National Natural Science Foundation of China under Grant No. 92470205 and 62222215, Beijing Municipal Science and Technology Project under Grant No. Z231100010323009, and Beijing Natural Science Foundation under Grant No. L233008. 
% Xin Zhao is the corresponding author.
\end{acks}

\balance
% \begin{acks}
% \end{acks}

\bibliographystyle{ACM-Reference-Format}
\bibliography{sample-base}


% \appendix
% \input{sec/sec-related}
% \input{sec/app}

\end{document}

%% file: sec/sec-intro.tex
\section{Introduction}

Despite advanced capabilities, large language models~(LLMs)~\cite{brown2020language, zhao2023survey} often struggle with knowledge-intensive challenges such as open-domain question answering (QA)~\cite{petroni2020kilt}. 
These limitations arise primarily from LLMs’ reliance on parametric knowledge, which often proves inadequate for real-time queries or domain-specific information, leading to factual hallucinations~\cite{cheng2024small}.
To migrate this issue, researchers have developed the retrieval-augmented generation~(RAG) technique~\cite{gao2023retrieval}, which first retrieves relevant information from an external knowledge base and then incorporates it as supplementary external knowledge into the input context. 
This approach not only enhances the model's parametric knowledge without additional training, but also expands LLMs' knowledge boundaries and improves their reliability and transparency~\cite{lewis2020retrieval}.
As a result, the potential of RAG has been harnessed in domains spanning complex tasks addressing~\cite{cheng2025think}, intelligent information assistants~\cite{lala2023paperqa}, and autonomous agents~\cite{ren2025holistically}.

Recently, researchers have shifted their focus towards understanding how RAG leverages knowledge, moving beyond merely aiming for task-specific performance improvements~\cite{wang2024rear}.
Existing studies have explored the impact of retrieval augmentation on the knowledge boundaries of LLMs and identified key factors that influence such knowledge boundaries in RAG scenario~\cite{ren2023investigating}. Additionally, effort examines the conflicts between parametric knowledge and non-parametric knowledge when LLM utilizes external context, proposing methods to prune conflicting modules and mitigate these conflicts~\cite{jin2024cutting}.
Despite recent progress, the field still lacks a comprehensive mechanistic understanding within the RAG framework of how LLMs navigate and reconcile both internal (parametric) and external (retrieved) knowledge~\cite{wang2024rear}, particularly at a compositional level.

Clear mechanistic insights are crucial for advancing our understanding of RAG's knowledge utilization and diagnosing potential pitfalls of RAG that emerged with transformer-based architectures~\cite{transformer}. Recent studies on model mechanisms have enhanced our understanding of the Transformer architecture~\cite{xiao2017attentional}. However, RAG introduces two unique challenges for interpretability, presenting new complexities compared to existing tasks like in-context learning (ICL) or mathematical reasoning~\cite{hanna2024does, dong2025longred}. First, RAG operates with unstructured data, which means the external knowledge inputs lack a fixed format. This factor increases the difficulty of analyzing how the model understands the inputs and extracting information as a reference. Second, RAG involves both internal and external sources of knowledge~\cite{jin2024tug}. This requires the LLM to integrate and select information from both parametric knowledge and contextual knowledge, demanding a nuanced understanding of how these sources interact.

In this study, we propose a comprehensive investigation of the intrinsic mechanisms governing LLMs within the RAG framework, systematically analyzing them from an interpretive perspective. 
Specially, we develop new analytical strategies that illuminate both macro-level knowledge streaming and micro-level module contributions within the Transformer architecture. 
Based on these perspectives, we propose two fundamental research questions: \textit{(1) How does knowledge stream within the RAG framework? (2) How do LLM modules function in knowledge utilization?}
To explore the first question, we quantitatively evaluate the stream of knowledge through each layer of the LLM using information flow methodologies. This approach allows us to track how internal and external knowledge streams evolve across layers within the RAG framework.
For the second question, we introduce a new approach to analyze the role of neurons, the multi-head attention (MHA) module, and the multi-layer perceptron (MLP) module in knowledge utilization. This enables us to conduct targeted interventions, modulating the LLM's reliance on internal versus external knowledge and providing deeper insight into the functionality of these modules.

We conduct a systematic study, focusing on two main aspects. First, we adopt a diverse range of perspectives, using various methods to explore, hypothesize, and validate specific mechanisms. Second, while uncovering the mechanisms, we also propose methods to leverage them for building controllable RAG systems.
 From these systematic investigations, we derive the following key insights:\\
(1) Knowledge streaming within the RAG framework can be identified into four distinct stages, regarding to knowledge utilization: knowledge refinement, knowledge elicitation, knowledge expression, and knowledge contestation.
Such a stage division can be re-examined by saliency analysis.\\
(2) The relevance degrees of retrieved passages direct the knowledge streaming in RAG. This relevance primarily affects the knowledge elicitation stage. This provides a fundamental explanation for how relevance discrepancies impact RAG's knowledge utilization and performance.\\
(3) We define a new metric, {K}nowledge {A}ctivation {P}robability {E}ntropy~({KAPE}), to identify neurons associated with internal and external knowledge. By specifically deactivating these neurons, we successfully altered the model's preference for the selection between two knowledge sources.\\
(4) We explore the contributions of MHA and MLP modules to knowledge generation. Our findings validate deep-layer knowledge competition and reveal that MLP layers play a role in verifying knowledge accuracy.\\

In summary, the contributions of this paper are as follows:

\begin{itemize}[leftmargin=1em, itemsep=0.5em, labelsep=0.5em]
    \item For the first time, we unveil the intrinsic knowledge utilization mechanisms of LLMs in RAG scenarios from two perspectives. At the macroscopic level, we examine the trends in knowledge streaming throughout the RAG process. At the microscopic level, we investigate the role of LLM modules in facilitating knowledge utilization within the RAG framework.
    \item From the knowledge streaming perspective, we observe four distinct stages in knowledge utilization of LLM-based RAG with saliency-based verification.
    Based on this, we identify the guiding role of relevance level between the query and external knowledge during the knowledge elicitation stage, and demonstrate that LLM evaluates such relevance through the information flow between them in this stage.
    \item From the perspective of LLM modules, we further investigate the functions of different modules.
    We propose a novel approach KAPE to identify knowledge-specific neurons within LLMs, and a deactivation mechanism to alter LLMs' expression tendencies of internal and external knowledge.
    Furthermore, we investigate the contributions of MLP and MHA to the formation of both knowledge.
\end{itemize}

%% file: sec/sec-preliminaries.tex
\section{Preliminaries}
\label{sec:preliminaries}
In this section, we provide a comprehensive description of the tasks involved, along with formal definitions of both internal and external knowledge. Furthermore, we elaborate on the specific experimental details that underpin our study.

\subsection{Task Description and Essential Definition}
\label{sec:definition}
The core concept of retrieval-augmented generation (RAG) is to enhance the model's generative capabilities by incorporating external information through a retrieval module. This broadens the model's knowledge and enables more accurate and contextually appropriate responses. In this study, we investigate the mechanisms of knowledge utilization in RAG within its typical application scenarios: open-domain question answering (ODQA)~\cite{chen2017reading}. The objective of the ODQA task is to extract relevant information~(passages in this study) from a vast external knowledge source to answer a specified query. The ODQA task covers a broad range of knowledge and allows large language models~(LLMs) to use both parametric knowledge (closed-book) and retrieval-augmented inputs. This makes it well-suited for studying how different types of knowledge are utilized in different settings.

\begin{table}[t]
\setlength\tabcolsep{2pt}
\centering
\small
\begin{tabular}{lcc}
\toprule
\textbf{Type} & \textbf{Acquisition Phase} & \textbf{Storage Mechanism} \\ 
\midrule
Internal & Training & Stored in LLM parameters (weights) \\ 
External & Inference & Provided as context (retrieved documents) \\ 
\bottomrule
\end{tabular}
\caption{Distinction between internal and external knowledge based on acquisition phase and storage mechanism.}
\vspace{-2ex}
\label{tab:defination}
\end{table}

In our study, we mainly focus on LLM backbones to conduct the empirical analysis. Referring to existing analyses on LLMs' knowledge~\cite{ren2023investigating}, we formally define the internal and external knowledge of the LLM based on the ODQA task. For \textit{internal knowledge}, given a question $q$ in natural language form, the answer of the question $a_\text{int}$ can be directly generated by the LLM with an instruction $I$:
\begin{equation}
    a_\text{int} = \text{LLM}(I, q).
    \label{eq:llm}
\end{equation}
Since the answer $a_\text{int}$ is generated solely from the LLM's internal parameters, it is considered a direct manifestation of the knowledge embedded within the LLM. This reflects the LLM's intrinsic capability to address knowledge-intensive tasks.
For \textit{external knowledge}, we enhance the LLM with a RAG approach. The instruction $I$ directs the LLM to extract an answer $a_\text{ext}$ to question $q$ using a selected passage subset $\mathcal{P}$ from the larger corpus $\mathcal{D}$ retrieved by the retriever $\text{R}$:
\begin{align}
    \mathcal{P} &= \text{Retriever}(\mathcal{D}), \\
    a_\text{ext} &= \text{LLM}(I, q, \mathcal{P}).
    \label{eq:retrieval}
\end{align}
In the RAG setting, the answer $a_\text{ext}$ generated by the LLM is regarded as an embodied manifestation of the external knowledge utilized during generation. 
Table~\ref{tab:defination} distinguishes internal and external knowledge by their acquisition phases and storage mechanisms.

In the following sections, we examine how RAG processes and utilizes internal and external knowledge. We analyze knowledge streams at the macroscopic level to understand overall trends. At the microscopic level, we examine the roles of neurons and modules in facilitating knowledge utilization. This helps unveil the mechanisms of RAG, paving the way for more reliable and controllable systems.

\subsection{Experimental Settings}
\label{sec:settings}

\paratitle{Evaluation Models.} \quad
To comprehensively investigate the knowledge streaming mechanisms of RAG, we analyze models from two popular open-source families: LLaMA and Qwen. For the LLaMA family, we include different versions and scales, specifically LLaMA-3-8B, LLaMA-3.1-8B and LLaMA-3-70B~\cite{dubey2024llama}. For the Qwen family, we study Qwen-2.5-1.5B and Qwen-2.5-7B~\cite{yang2024qwen2}. All selected models are Base versions.

\paratitle{Dataset.} \quad
We collect three extensively adopted open-domain QA benchmark datasets.
including two single-hop QA datasets~(\textit{Natural Questions}~\citep{nq}, \textit{TriviaQA}~\citep{joshi2017triviaqa}) and a multi-hop QA dataset~(\textit{HotpotQA}~\citep{yang2018hotpotqa}). 
Natural Questions~(NQ) is a dataset of question-answer pairs derived from real Google search queries, with answers annotated by human experts. 
TriviaQA contains trivia questions paired with annotated answers and supporting evidence documents. 
HotpotQA features question-answer pairs that require multi-hop reasoning to determine the correct answer.

\paratitle{Retrieval Augmentation.} \quad
As dense text retrieval is demonstrated effective in many scenarios~\cite{pair}, we utilize the open-sourced RocketQAv2~\cite{rocketqav2} as the passage retriever and use the Wikipedia dump, the same as the previous work~\citep{rocketqa}, as the external knowledge corpus. For each question, we recall the most relevant passages and filter only one \textit{gold passage} that has a low overlap degree with internal knowledge $a_\text{int}$ and contains the correct answer, serving as the external knowledge source for the LLM. Furthermore, to ensure rigorous analysis, we constructed an additional \textit{fake passage} based on the gold passage for each query, where the correct answer is replaced with the incorrect one. The advantage of this approach lies in the fact that the fake passages serve as a knowledge source completely unseen by the LLM. This guarantees no overlap between the LLM's internal knowledge and the external knowledge within the fake passages, enabling more precise phenomenons of the external knowledge.

\paratitle{Implementation Details.} \quad 
Our experimental setup is structured into two main phases: first, the extraction of embodied manifestations of external knowledge as discussed in Section~\ref{sec:definition}, and second, the analysis of corresponding methods described in Section~\ref{sec:flow-anas} and Section~\ref{sec:deeper}.
Following previous work~\cite{tang2024dawn, tang2024unleashing}, all responses were generated using greedy decoding.
All experiments are conducted on 8 NVIDIA A100 GPUs with 80GB of memory, using bfloat16 precision. 

%% file: sec/sec-analysis.tex
\section{Knowledge Streams within RAG}
\label{sec:flow-anas}

In this chapter, we analyze the knowledge utilization process of LLMs in RAG scenarios from a macroscopic perspective. Specially, we examine how internal and external knowledge are utilized as the LLM layers deepen, via a knowledge stream perspective. We begin by observing the knowledge streaming process using attention and saliency analysis. Additionally, we explore how the relevance of external knowledge impacts the dynamics of knowledge streaming.

\subsection{Information Flow Methodology}

In this part, we first establish the relationship between knowledge streaming and information flow. We then introduce the two information flow analysis methodologies and the metrics we used. 

\paratitle{Knowledge Streaming and Information Flow.}  \quad In RAG systems, knowledge streaming is the dynamic integration of external knowledge into a model's processes. This occurs through token interactions within the Transformer architecture~\cite{transformer, wang2023label, tang2025unlocking}. Tokens representing retrieved knowledge impact response generation and modify other tokens' states. We observe these patterns to trace knowledge manipulation within the model. Our research uses two main methods to analyze these patterns: attention-based and saliency-based information flow analyses. These methods help us explore how information flow varies across layers, showcasing the knowledge streaming process.

\paratitle{Attention-based Information Flow.}  \quad
Attention scores in the Transformer architecture directly reflect the flow of information~\cite{bastings2020elephant, jacovi2020towards}. The dot product's score between a target token's query vector and a source token's value vector determines the influence extent that the source token influences the target token's hidden state. By analyzing these attention scores across different layers, we can assess the direct information flow within the LLM, providing insights into how information is dynamically managed and utilized.
To facilitate the analysis, we first define three components within the RAG input instructions:
\begin{itemize}[leftmargin=1em, itemsep=0.5em, labelsep=0.5em]
    \item \textbf{C}~({context}): the passages utilized for retrieval augmentation.
    \item \textbf{K}~(\textit{key}): the potential answers extracted by the LLM from the context, obtained following Equation~(\ref{eq:retrieval}).
    \item \textbf{Q}~(\textit{query}): the question to be answered.
    \item \textbf{A}~(\textit{answer prompt}): the guiding message at the end of the instruction that directs the model to generate the answer.
\end{itemize}
Based on this, we further define three quantitative metrics to evaluate the information flow based on attention scores:
\begin{itemize}[leftmargin=1em, itemsep=0.5em, labelsep=0.5em]
    \item \(\text{IF}_a^{kc}\): the attention information flows from key tokens to the context tokens.
    \item \(\text{IF}_a^{kq}\): the attention information flows from key tokens to query tokens.
    \item \(\text{IF}_a^{ka}\): the attention information flows from key tokens to answer tokens.
\end{itemize}

The three metrics are calculated based on attention matrices adopting the following equation:
\begin{equation}    
    IF_{a}^{XY}=\sum_h\sum_i\sum_j {\bm{A_{h,l}}(i, j)}, \quad i\neq j, i\in X, j\in Y,
    \label{eq:if}
\end{equation}
where X and Y denote different component tokens defined above, \(\bm{ A_{h,l}} \) represents the value of the attention matrix corresponding to the \( h \)-th attention head in the \( l \)-th layer.

\paratitle{Saliency-based Information Flow.} \quad
Moreover, following previous studies on interpreting Transformer mechanisms~\cite{wang2023label}, we further analyze the information flow in RAG using the gradient-based saliency method~\cite{bastings2020elephant}. This method evaluates the marginal effect of specific inputs or parameters through gradients. Formally, we compute the saliency matrix using attention metrics and Taylor expansion~\cite{michel2019sixteen}:
\begin{equation}
    S_l = \sum_h \big| \nabla_{\bm{A_{h,l}}} \mathcal{L_\text{sft}}(x) \odot \bm{A_{h,l}^{\mathrm{T}}} \big| ,
\end{equation}
where $\odot$ denotes the Hadamard product, \( x \) is the input and \( \mathcal{L}_\text{sft}(x) \) represents the supervised fine-tuning loss.
The saliency matrix $S_l$ measures the significance of information flow across tokens.
As with attention-based information flow metrics, we define three quantitative metrics \(\text{IF}_s^{kc}\), \(\text{IF}_s^{kq}\) and \(\text{IF}_s^{ka}\) to evaluate the information flow based on saliency scores using the same calculation format with Equation~(\ref{eq:if}).
This method offers deeper insights into how each input token affects the final prediction by accounting for the entire gradient path, from input embeddings to output, rather than just a single attention step. For Transformer-based LLMs, it captures token-level influence trends under specific optimization objectives.

Using these two information flow methodologies, we analyze the knowledge streaming process in RAG, uncovering its phased trends. This provides a structured perspective for understanding the dynamics of knowledge streaming, as discussed in the remainder of this section.

\begin{figure}
    \centering
    \includegraphics[width=1\linewidth]{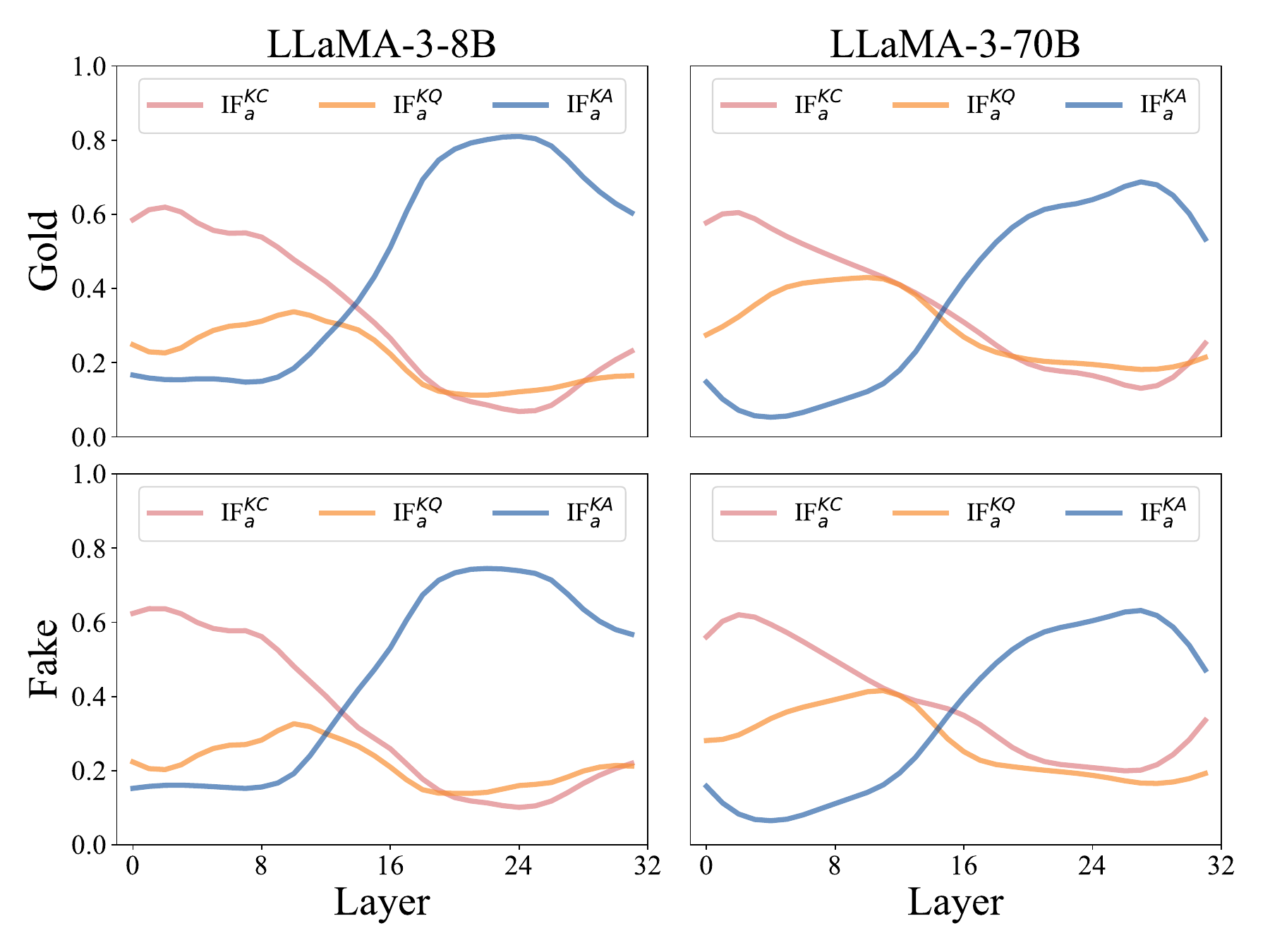}
    \caption{Attention-based information flows across various directions in two versions of LLaMA-3 with different parameter scales, evaluated under the RAG setting using gold and fake passages
    on the average of three datasets.
    }
    \label{fig:att}
\end{figure}

\subsection{Knowledge Streaming in RAG}
\label{sec:att_if}
In this part, we employ the attention-based information flow approach to depict the dynamics of knowledge transfer within RAG. Specifically, we quantify the knowledge streaming process using three evaluation metrics \(\text{IF}_a^{kc}\), \(\text{IF}_a^{kq}\) and \(\text{IF}_a^{ka}\).

\subsubsection{Experimental Results and Analysis} 

Figure~\ref{fig:att} illustrates the variations in RAG information flows across different LLM layers within LLMs based on gold passages~(defined in Section~\ref{sec:settings}), we conduct the evaluation on four widely used open-sourced LLMs. Here we will analyze the trends for the three information flows separately, examine the differences between models, and compare the effects of different augmented documents.

\paratitle{Key-to-Context Information Flow ~(\(\text{IF}_a^{kc}\)).} \quad
This flow shows a steady decline with a slight increase in the deeper layers. In the early layers, the interaction between the key and context is strong. This enables the LLM to quickly refine and adjust its understanding of the context. As layers deepen, this influence weakens, indicating that the context information becomes stable and less reliant on further key-context interactions.

\paratitle{Key-to-Query Information Flow ~(\(\text{IF}_a^{kq}\)).} \quad
Initially, the flow increases before decreasing in the deeper layers. This suggests that as the LLM processes external knowledge, it gradually integrates this information into the hidden states of the query. In the deeper layers, the influence reduces, indicating that the knowledge has been sufficiently encoded by the LLM at this stage.

\paratitle{Key-to-Answer Information Flow ~(\(\text{IF}_a^{ka}\)).} \quad
This flow begins to increase around the first quarter of the layers before gradually decreasing in the last quarter. This suggests that external knowledge contributes to the answer-generation process in the middle layers, improving the LLM's ability to form a response. In the deeper layers, the influence reduces as the LLM consolidates the answer information with less reliance on external input.

\begin{figure}
    \centering
    \includegraphics[width=0.95\linewidth]{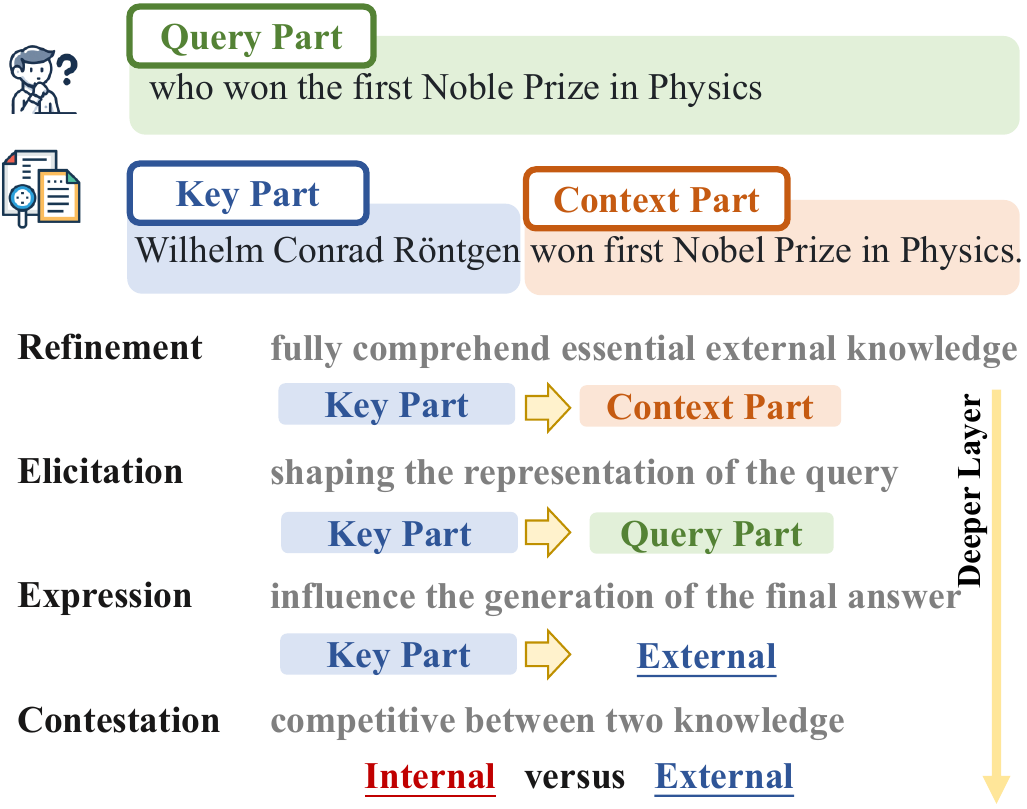}
    \caption{Illustrations of the four stages of knowledge streaming within RAG.}
    \label{fig:stage}
\end{figure}

\paratitle{Consistency Across LLMs.} \quad
We observe consistent results across various LLM families and scales, demonstrating the generalizability of our findings. A similar trend is also evident in LLaMA-3.1-8B and Qwen-2.5-7B. Due to space limitations, we have to leave out the results of the two LLMs.

\paratitle{Comparison of Gold and Fake Passages.} \quad 
We further compare the effect of using gold versus fake passages on RAG information flow. In the case of fake passages, the correct answers~(\textit{key}) in gold passages are replaced with incorrect alternatives~(defined in Section~\ref{sec:settings}). This not only alters the \textit{key part} but also ensures the content is inconsistent with the LLM's pretraining data, eliminating interference from internal knowledge. Despite these changes, the information flow patterns in fake passages remained largely consistent with those from gold passages. This highlights that the LLM's internal processing remains stable, even when the external key is incorrect, further validating the robustness of the findings.

\begin{figure*}
    \centering
    \includegraphics[width=0.99\linewidth]{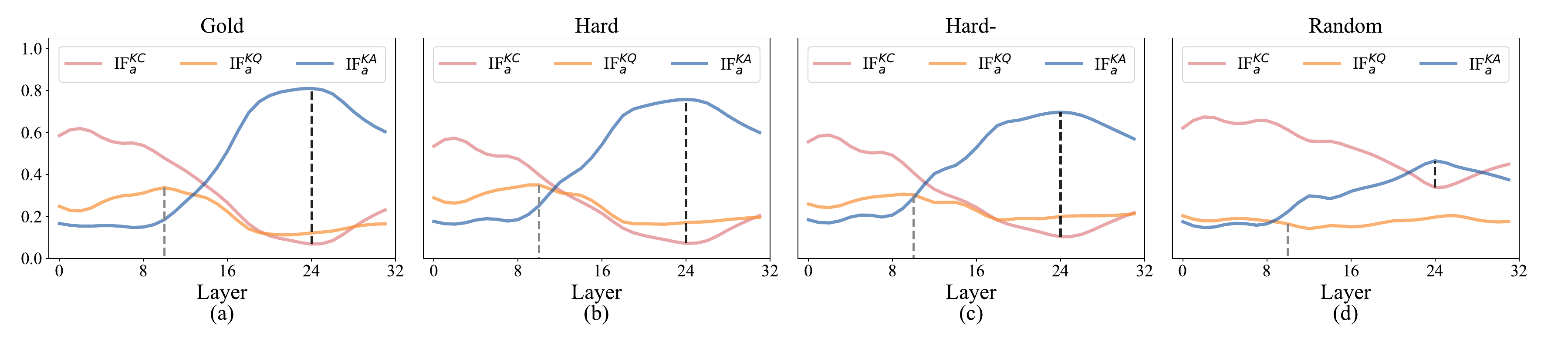}
    \caption{Attention-based information flow based on external passages of various relevance with the query.
    }
    \vspace{-2ex}
    \label{fig:relevance}
\end{figure*}

\subsubsection{Knowledge Streaming Exhibits Multi-stage Nature in RAG}
\label{sec:phase}
By observing the knowledge streaming, we propose a hypothesis that knowledge traverses multiple streaming stages in RAG.
Specifically, we delineate four distinct stages:

\begin{itemize}[leftmargin=1em, itemsep=0.5em, labelsep=0.5em]
    \item \textbf{Stage 1: Knowledge Refinement.} During this phase, external knowledge, represented by the interaction between the context and key, is deeply integrated, enabling the LLM to fully comprehend the context and distill essential external knowledge.

    \item \textbf{Stage 2: Knowledge Elicitation.} As the context is absorbed, the flow of knowledge from the key to the query intensifies, transmitting the refined contextual information and shaping the representation of the query.

    \item \textbf{Stage 3: Knowledge Expression.} With the continued flow between external knowledge and the query, external knowledge begins to significantly influence the generation of the final answer.

    \item \textbf{Stage 4: Knowledge Contestation.} A competitive dynamic emerges between external and internal knowledge within the LLM, ultimately determining the final answer.
\end{itemize}

Figure~\ref{fig:stage} illustrates the four stages of knowledge streaming within RAG. Each stage is defined by distinct directions of knowledge streams, supporting effective knowledge utilization. Note that these stages are not strictly discrete, as transitional layers may exist between stages.

\subsection{Corroboration Analysis on Saliency}

\begin{figure}
    \centering
    \includegraphics[width=1\linewidth]{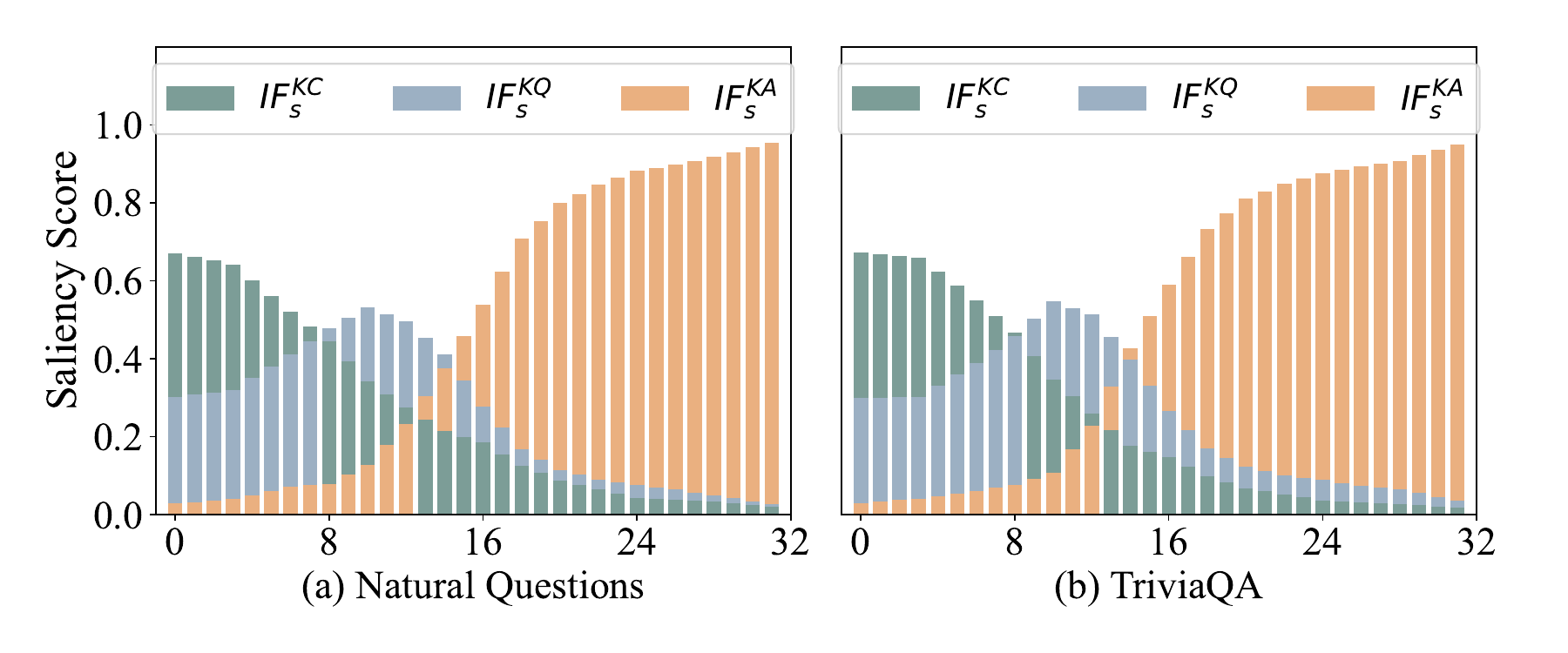}
    \caption{Saliency score on three information flow directions on Natural Questions and TriviaQA dataset for LLaMA-3-8B.}
    \vspace{-2ex}
    \label{fig:saliency}
\end{figure}

To verify the rationality of the proposed four-stage RAG knowledge streaming hypothesis, we conduct corroboration from the perspective of saliency-based information flow. As introduced in the section, the saliency score represents the changing trend of attention-based information flow, reflecting whether the LLM aims to enhance or reduce the expression of knowledge at a given moment.
Similar to Section~\ref{sec:att_if}, we quantify the knowledge streaming process using three saliency-based information flow metrics \(\text{IF}_s^{kc}\), \(\text{IF}_s^{kq}\) and \(\text{IF}_s^{ka}\), 
Figure~\ref{fig:saliency} illustrates the saliency-based metrics that vary in different layers evaluated on LLaMA-3-8B.

First, we observed \textbf{a trend of modifying knowledge streaming similar to the one discussed in Section~\ref{sec:att_if}} on knowledge streaming. During the knowledge refinement stage, the LLM tends to increase the flows from the context to the key, refining external knowledge. In the knowledge elicitation stage, the LLM enhances the flows from the key to the query. In the knowledge expression stage, the LLM demonstrates a pronounced increase in its tendency to enhance the flows from the key to the answer prompt. This trend remains steady and consistent during the knowledge contestation stage.

Moreover, the knowledge streaming patterns shown in Figure~\ref{fig:att} present that the flows from the key to the answer decreases in the final stage. However, the LLM tends to enhance it. \textbf{This suggests that the decline in external knowledge streaming may be influenced by the LLM’s internal knowledge streaming.} The difference confirms that a contestation relationship indeed exists between external and internal knowledge during the final stage.

In summary, these findings support the segmentation of knowledge streaming stages in the RAG process from a different perspective.

\subsection{Relevance Guides the Knowledge Streaming}
The previous study has demonstrated that the relevance of external passages significantly influences the performance of LLMs in RAG scenarios~\cite{ren2023investigating}. Based on this, we have strong reason to hypothesize that the influence of the given passage relevance on RAG performance primarily stems from its impact on the knowledge streaming within the model. In this part, we explore the dynamics of knowledge streaming in RAG under varying settings of passage relevance with queries.

\subsubsection{External Knowledge Source with Different Relevance.}
Given an input question, we follow the previous work~\cite{ren2023investigating}, and categorize passages with varying degrees of relevance as follows:
\begin{itemize}[leftmargin=1em, itemsep=0.5em, labelsep=0.5em]
    \item \textbf{Positive}~(\textit{positive passage}): the passage that contains the correct answer, as defined in Section~\ref{sec:settings}.
    \item \textbf{Hard}~(\textit{hard negative passage}): the passage relevant to the query but lacking the correct answer, sampled from the top-ranked retrieval results for the query.
    \item \textbf{Hard-}~(\textit{less hard negative passage}): the passage weakly relevant to the query but lacking the correct answer, randomly sampled from the retrieval results for the query.
    \item \textbf{Random}~(\textit{randomly sampled passage}): the passage irrelevant to the query and devoid of the correct answer, randomly sampled from the entire corpus.
\end{itemize}
Based on the passages with varying degrees of relevance to the question, we can conduct a comprehensive evaluation of the influence of external knowledge relevance on the knowledge streaming within RAG.

\subsubsection{Relevance Guides the Knowledge Streaming for Various External Knowledge Relevance} 
We firstly examine how modifications in the passage's relevance affect the RAG knowledge streaming, Figure~\ref{fig:relevance} illustrates the results of attention-based information flow.

Firstly, we observe that when processing passages with varying degrees of relevance, \textbf{the overall trend of knowledge streaming remains consistent}. The four stages we previously defined are clearly identifiable across all three scenarios with different relevance levels~(first three sub-figures in Figure~\ref{fig:relevance}).

Secondly, \textbf{as relevance decreases, the changes in the three information flows and their differences become less noticeable}. For more relevant passages, external knowledge streaming increases, indicating that the LLM uses these passages more extensively to generate responses.

Specifically, the gray dashed line in the figure highlights the changes in~\(\text{IF}_a^{kq}\) during the knowledge elicitation stage. This shows that the LLM’s integration of external knowledge into query understanding weakens significantly with decreasing relevance. The gray dashed line also marks the difference between the key-to-answer~\(\text{IF}_a^{ka}\) and key-to-context~\(\text{IF}_a^{kc}\) flows during the knowledge expression stage. \textbf{We can observe that LLM minimizes its use of external knowledge when handling irrelevant content}.

Overall, these findings provide a foundational explanation for existing work~\cite{ren2023investigating} and demonstrate the LLM’s distinctive characteristic to utilizing external knowledge depending on external knowledge relevance.

\begin{figure}
    \centering
    \includegraphics[width=0.8\linewidth]{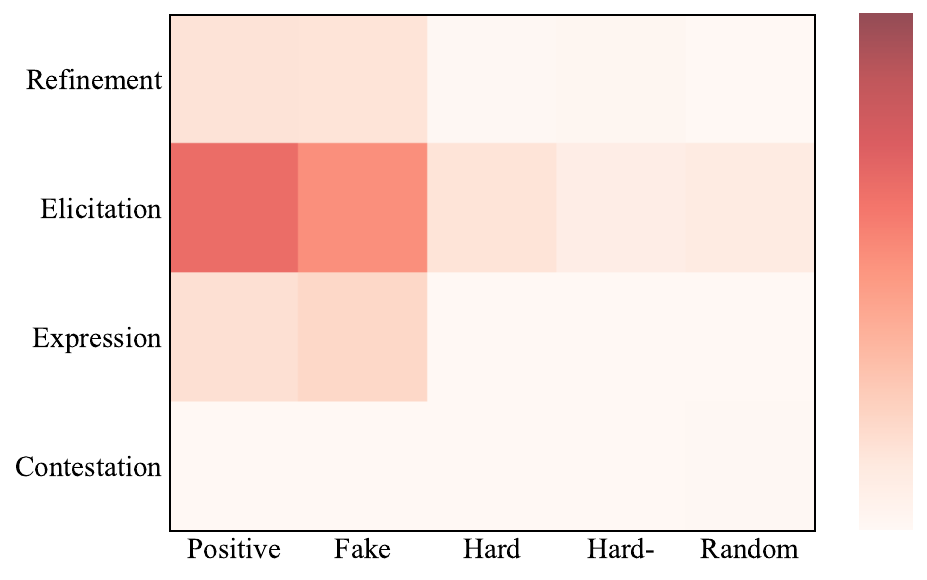}
    \caption{The figure shows the decrease in the probability of generating internal knowledge. This occurs after disrupting the key-to-query knowledge streaming across the four knowledge streaming stages.}
    \label{fig:heat}
\end{figure}

\subsubsection{Knowledge Elicitation with Relevance Assessment}

Previous results indicate that the influence of external knowledge relevance on LLM knowledge streaming emerges relatively early during the knowledge elicitation stage~(gray dashed line in Figure~\ref{fig:relevance}). We hypothesize that during this stage, the interaction between external knowledge and the query enables the LLM to perform relevance discrimination, and it further guides LLM on whether to incorporate external knowledge. To validate this hypothesis, we conducted experiments by supplying passages with varying levels of relevance, and cut off the knowledge streaming between key and query at different stages. We then measured the differences in the final layer probabilities for generating external knowledge:

\begin{equation}
    d = p ( a_\text{ext} | q, p, \bm{M_{\text{causal}}}) - p ( a_\text{ext} | q, p, \bm{M_{\text{causal}}} + \bm{M_{\text{k2q}, s}} ) ,
\end{equation}
where $s$ denotes the interrupted knowledge streaming stage, $\bm{M_{\text{causal}}}$ is the causal mask for decoder-only LLMs, and $\bm{M_{\text{k2q}, s}}$ represents the mask of the attentions from key part to query part. 
Figure~\ref{fig:heat} presents the results with a heatmap. It is evident that upon cutting the knowledge streaming from key to query during the knowledge elicitation stage, the  increase in the expression of external knowledge is more pronounced for passages with higher similarity. This finding supports our hypothesis that relevance judging is performed during the knowledge elicitation stage of RAG.

%% file: sec/sec-indepth.tex
\section{Modules Function in  Knowledge Utilization}
\label{sec:deeper}

In this chapter, we study knowledge utilization in LLMs within RAG scenarios from a microscopic perspective, focusing on the roles of different modules. Specifically, we first analyze the role of neurons and their activations in the utilization of internal and external knowledge. Furthermore, we investigate the respective functions of the MLP and MHA modules in the expression of knowledge.

\subsection{Affects of Knowledge-Specific Neurons}
\label{sec:kape}

Our findings reveal distinct patterns across layers, suggesting that, similar to the human brain, LLMs have specialized regions and neurons. The utilization of internal and external knowledge appears to rely on different neurons.
In this part, we propose a method to identify neurons that activate different types of knowledge~(Section~\ref{sec:identify}). By deactivating these neurons, we aim to change the model’s preference for internal or external knowledge~(Section~\ref{sec:deact}).

\subsubsection{Identification of Knowledge-specific Neurons.} 
\label{sec:identify}
In this part, we introduce a method to identify neurons that manage internal and external knowledge. We first describe the notion of neuron activation, followed by our proposed methodology for identifying knowledge-specific neurons.

\paratitle{Neuron Activation and Gating in LLMs.} \quad
For recent LLMs such as LLaMA, the MLP layers use a gating mechanism with Gated Linear Units~(GLU)\cite{shazeer2020glu}. The MLP layer is defined as:
\begin{equation}
    \text{MLP}(\bm{h}) = \left(\text{act\_fn}(\bm{h} \bm{W}_{\text{gate}} \odot \bm{h} \bm{W}_{\text{up}})\right) \bm{W}_{\text{down}}.
\end{equation}

Consistent with previous work~\cite{nair2010rectified}, we define that the $j$-th neuron inside the $i$-th feedforward network (FFN) layer is considered to be activated if its respective activation values from $\text{act\_fn}(\bm{h}_i \bm{W}_{\text{gate}, i})_j$ exceed zero.
Formally, for the $j$-th neuron in the $i$-th layer of the model, we first compute its activation probability when utilizing both internal and external knowledge:

\begin{equation}
p^k_{i,j} = \left\| \mathbb{E}_k \bigg[ \Theta \big( \text{act\_fn} \big( \bm{h}^i \bm{W}^i_j \big) \big) \bigg] \right\|_1^\text{norm},
\label{eq:actprob}
\end{equation}
where $\Theta(\cdot)$ denotes the {Heaviside step function}. Using L1 normalization, we transform the raw activation probabilities of neurons across different data samples into a distribution reflecting internal and external knowledge.
This process effectively maps the original probabilities to a distribution focused on the types of knowledge, enabling us to better isolate and analyze the neurons associated with each knowledge type.

\paratitle{Quantifying Neuron Knowledge Preference via KAPE.} \quad 
We identify neurons that handle internal and external knowledge. Inspired by the existing study~\cite{tang-etal-2024-language}, we define \textbf{K}nowledge \textbf{A}ctivation \textbf{P}robability \textbf{E}ntropy~(KAPE). Using the L1-normalized probabilities calculated from Equation~\ref{eq:actprob}, we calculate the probabilities $p_{i,j}^{\text{IK}}$ and $p_{i,j}^{\text{EK}}$ for internal and external knowledge respectively. The KAPE for each neuron is computed as follows: 
\begin{equation} \text{KAPE}_{i,j} = - \left( p_{i,j}^{\text{IK}} \log(p_{i,j}^{\text{IK}}) + p_{i,j}^{\text{EK}} \log(p_{i,j}^{\text{EK}}) \right), 
\end{equation}
where a low KAPE score suggests that the $j$-th neuron in the $i$-th layer has a high activation probability for one type of knowledge and a low probability for another, making it a knowledge-specific neuron. 

\paratitle{Implementation Details.} \quad
The identification process consists of two main steps: calculating the KAPE scores and refining the neurons. First, we compute the KAPE scores for each neuron based on their activation probabilities under both RAG instructions with gold passages and a closed-book setting, following L1 normalization to derive \(p_{i,j}^{\text{IK}}\) and \(p_{i,j}^{\text{EK}}\) in Equation~\ref{eq:actprob}, and calculated the \(\text{KAPE}_{i,j}\) score for each neuron. Second, we select the top 1\% of neurons with the lowest KAPE scores as knowledge-specific. We then apply a threshold to retain only those neurons with significant activation probabilities in either RAG or closed-book scenarios. These steps help identify neurons specific to internal or external knowledge.

\begin{table*}[t]
\centering
\small
\begin{tabular}{lccccccccccccc}
\toprule
\multicolumn{1}{c}{\multirow{2.5}{*}{\textbf{Document}}} & \multicolumn{1}{c}{\multirow{2.5}{*}{\textbf{Setting}}} & \multicolumn{3}{c}{{Natural Questions}} & \multicolumn{3}{c}{{TriviaQA}} & \multicolumn{3}{c}{{HotpotQA}} & \multicolumn{3}{c}{{Average}} \\
\cmidrule(r){3-5} \cmidrule(r){6-8} \cmidrule(r){9-11} \cmidrule(r){12-14}
\multicolumn{1}{c}{} & \multicolumn{1}{c}{} & \textbf{EM} & \textbf{CEM} & \textbf{F1} & \textbf{EM} & \textbf{CEM} & \textbf{F1} & \textbf{EM} & \textbf{CEM} & \textbf{F1} & \textbf{EM} & \textbf{CEM} & \textbf{F1} \\

\midrule
\multicolumn{14}{c}{\textit{LLaMA-3-8B}} \\
\midrule
Closed-Book & Normal & 22.41 & 28.45 & 33.53 & 58.62 & 63.22 & 62.70 & 16.95 & 19.21 & 25.29 & 32.66 & 36.96 & 40.51 \\
\midrule
\multirow{3}{*}{Gold Document} & Normal & 42.24 & 49.14 & 56.19 & 78.74 & 85.06 & 83.64 & 46.89 & 55.93 & 59.57 & 55.96 & 63.38 & 66.47 \\
\multirow{3}{*}{} & Deactivate \textit{IK} & 39.66 & 43.10 & 52.49 & 68.97 & 75.29 & 77.39 & 39.55 & 48.02 & 51.81  & 49.39(-6.57) & 55.47(-7.91) & 60.56(-5.91) \\
\multirow{3}{*}{} & Deactivate \textit{EK} & 37.93 & 42.24 & 52.82 & 63.22 & 65.52 & 70.97   & 28.81 & 36.16 & 41.55 & 43.32(-12.64) & 47.97(-15.41) & 55.11(-11.36)  \\
\midrule
\multirow{3}{*}{Noisy Document} & Normal & 9.48 & 12.07 & 15.37 & 35.06 & 39.08 & 40.20 & 5.71 & 8.00 & 14.08 & 16.75 & 19.72 & 23.22 \\
\multirow{3}{*}{} & Deactivate \textit{IK} & 6.03 & 9.48 & 12.86 & 30.46 & 35.63 & 35.46 & 3.43 & 4.57 & 10.92 & 13.31(-3.44) & 16.56(-3.16) & 19.75(-3.47) \\
\multirow{3}{*}{} & Deactivate \textit{EK} & 10.93 & 15.52 & 17.75 & 36.78 & 37.93 & 39.73 & 6.43 & 8.57 & 14.52 & 18.05(+1.30) & 20.67(+0.95) & 24.00(+0.78) \\
\midrule
% \midrule
\multicolumn{14}{c}{\textit{Qwen-2.5-1.5B}} \\
\midrule
Closed-Book & Normal  & 13.39 & 18.75 & 21.68 & 21.30 & 22.22 & 23.18 & 7.65 & 9.41 & 14.57 & 14.11 & 16.79 & 19.81 \\
\midrule
\multirow{3}{*}{Gold Document} & Normal  & 41.07 & 41.07 & 51.11 & 73.15 & 73.15 & 77.74 & 36.47 & 41.76 & 49.88 & 50.23 & 51.99 & 59.58\\
\multirow{3}{*}{} & Deactivate \textit{IK} & 39.29 & 40.18 & 49.83 & 70.37 & 70.37 & 73.77 & 26.47 & 30.00 & 37.15 & 45.38(-4.85) & 46.85(-5.14) & 53.58(-6.00)\\
\multirow{3}{*}{} & Deactivate \textit{EK} & 37.50 & 39.29 & 49.16 & 68.52 & 69.44 & 72.93 & 18.82 & 21.76 & 27.43 & 41.61(-8.62) & 43.50(-8.49) & 49.84(-9.74)\\
\midrule
\multirow{3}{*}{Noisy Document} & Normal & 4.46 & 5.36 & 11.13 & 12.04 & 12.04 & 13.33 & 2.01 & 2.01 & 6.75 & 6.17 & 6.47 & 10.40 \\
\multirow{3}{*}{} & Deactivate \textit{IK} & 2.68 & 3.57 & 8.23 & 10.19 & 10.19 & 10.19 & 1.51 & 2.51 & 4.53 & 4.79(-1.38) & 5.42(-1.05) & 7.65(-2.75)
 \\
\multirow{3}{*}{} & Deactivate \textit{EK} & 4.46 & 7.04 & 11.98 & 15.44 & 12.37 & 14.35 & 2.51 & 4.07 & 5.31 & 8.80(+1.63) & 7.83(+1.36) & 10.55(+0.15)\\
\bottomrule
\end{tabular}
\caption{Knowledge-specific neuron deactivation results on three QA datasets: Natural Questions, TriviaQA, and HotpotQA, using two LLMs~(LLaMA-3-8B and Qwen-2.5-1.5B) across different document settings. Metrics include EM~\cite{lee2019latent}, CEM~\cite{mallen2023not}, and F1 scores~\cite{karpukhin2020dense}, with average results reported. Results reflect the effect of deactivating IK and EK Neurons. }
\vspace{-2ex}
\label{tab:main-result}
\end{table*}

\subsubsection{Knowledge Guiding with Knowledge-Specific Neuron}
\label{sec:deact}
After identifying the relevant neurons, we can verify the accuracy of our selection and their functional roles by deactivating the neurons associated with both internal and external knowledge.

\paratitle{Neuron Deactivation Experiments.} \quad
To conduct the deactivation experiment, we first randomly selected retrieval results. We then measured the change in perplexity during the generation process when different neurons were deactivated. Specifically, we set the activation values of internal and external knowledge neurons to zero, and calculated the perplexity of generating internal and external knowledge separately. We identify the knowledge-specific neurons on the Natural Questions~\cite{nq} dataset, and further conduct our experiments on both single-hop QA datasets~(Natural Questions and TriviaQA~\cite{joshi2017triviaqa}) and multi-hop QA dataset~(HotpotQA~\cite{yang2018hotpotqa}).

\paratitle{Main Results.} \quad
The results illustrating the impact of neuron activation manipulation on knowledge utilization are shown in Table~\ref{tab:main-result}. It can be observed that:
\begin{itemize}[leftmargin=1em, itemsep=0.5em, labelsep=0.5em]
\item 
\textbf{Firstly, the results demonstrate that deactivating knowledgespecific neurons effectively controls the expression of internal and external knowledge.} Specifically, with gold documents, deactivating external knowledge neurons causes a greater decline in factual accuracy compared to internal knowledge neurons. In contrast, with noisy documents, deactivating external knowledge neurons reduces the influence of irrelevant content, whereas deactivating internal knowledge neurons makes the output less accurate.
\item 
\textbf{Secondly, our experiments confirm the generalizability of our approach.} Knowledge-specific neurons identified on Natural Questions were tested on other datasets, including TriviaQA and HotpotQA. The results show consistent effectiveness across single-hop and multi-hop tasks. We also observed similar knowledge utilization trends on all datasets. Furthermore, experiments with LLaMA and Qwen models showed consistent patterns, proving the method works across different model families.
\item 
\textbf{Thirdly, the impact of neuron editing on factual accuracy is influenced by question difficulty and retrieval quality.} When questions are less challenging, deactivating internal neurons has a greater effect on performance. Conversely, when the quality of retrieved documents is high, deactivating external neurons results in lower accuracy.
\end{itemize} 
These findings demonstrate that manipulating neuron activation shifts the LLM's reliance between internal and external knowledge, providing valuable insights for building controllable RAG systems.

\begin{figure}
    \centering
    \includegraphics[width=0.82\linewidth]{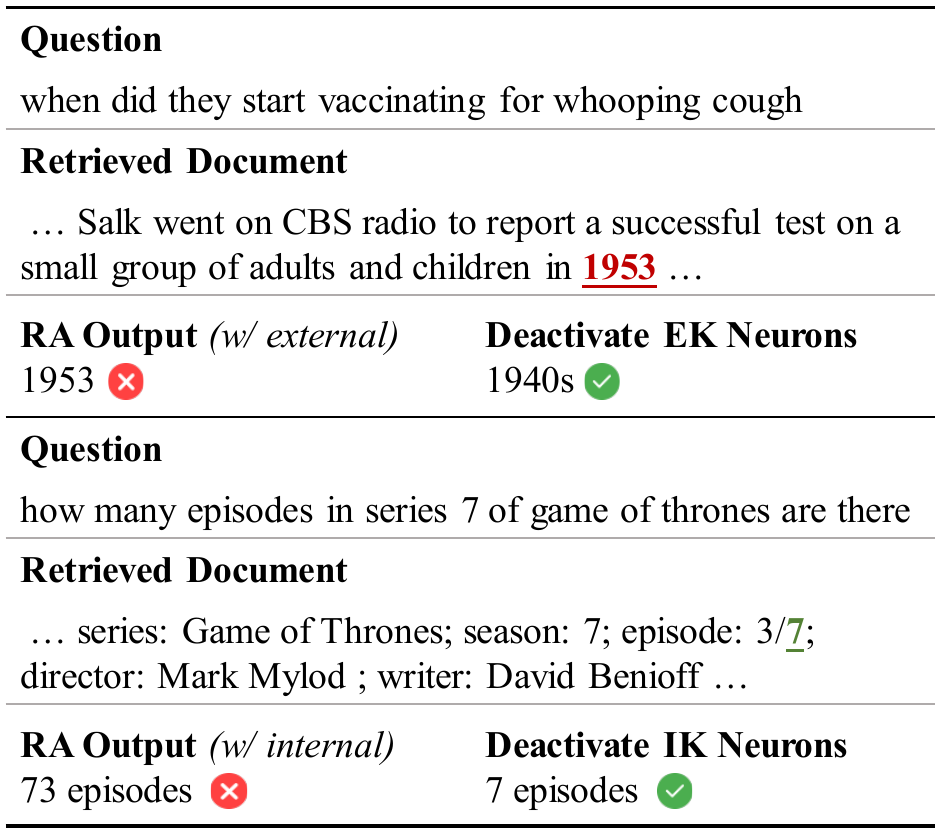}
    \caption{Cases for the LLM's knowledge utilization tendencies using knowledge-specific neuron deactivation.}
    \label{fig:case}
    \vspace{-2ex}
\end{figure}

\paratitle{Case Study.} \quad
We further demonstrate how knowledge-specific neuron deactivation impacts the tendency of knowledge utilization with two cases, as shown in Figure~\ref{fig:case}.
In the first case. The passage provided contains an incorrect answer. However, the LLM is initially misled by it, maybe for the reason that the passage is relevant. By deactivating neurons associated with external knowledge, the LLM's dependence shifts towards its internal knowledge, which correctly adjusts the response.
In the second case, the external passage correctly answers the query, but the LLM fails to leverage this external information at first. Deactivating neurons linked to internal knowledge enhances the LLM’s ability to utilize external data, leading to a successful integration of the correct answer.

\subsection{LLM modules Contribute to Internal and External Knowledge Formation}

In this part, we further analyze the roles of the two main modules in Transformers in knowledge utilization. Specifically, we examine how MHA and MLP influence the selection between internal and external knowledge. Additionally, we explore how the LLM verifies the consistency between its internal knowledge and external information, and how it handles noisy information with factual errors.

\subsubsection{Analysis Methodologies}
\label{sec:residual}
To study from a modular perspective, we first introduce methodologies for residual steam and early decoding analysis. Using these approaches, we can delve deeper into the influence of each module on the final answer generation.

\paratitle{Residual Stream.} \quad
To analyze the contribution of each module within Transformer-based models, we consider them as a series of residual stream~\cite{elhage2021mathematical, lv2024interpreting}. Each module in these models adds newly processed information into the stream through residual connections, and the sum of these contributions forms the final output. This setup isolates each module's impact on the hidden state. It reveals their roles in generating the final answer.

\paratitle{Early Decoding.} \quad
For interpretability, the early decoding strategy projects each module's incremental updates onto a human-readable vocabulary space~\cite{lv2024interpreting}. Specifically, before layer normalization and linear transformation, the updates are mapped as follows: 
\begin{equation}
    [\text{logit}_1^{l}, ..., \text{logit}_{\bm{|V|}}^{l}] = \bm{W^U} \cdot \text{LayerNorm}(\bm{h}_l),
\end{equation}
where $\bm{h_{l}} \in \mathbb{R}^{d}$
denotes the hidden state in the layer $l$, $\bm{W^U} \in \mathbb{R}^{|V| \times d}$ denotes the unembedding metric, LayerNorm denotes the pre-unembedding layer normalization, and $V$ denotes the token vocabulary of the model.
This mapping generates logits at intermediate layer $l$, serving as an early-exit mechanism~\cite{teerapittayanon2016branchynet}. These logits measure the model's predictions at each layer, offering insight into the contribution of individual modules to the output.

\begin{figure}[t]
	\centering
	\includegraphics[width=0.45\textwidth]{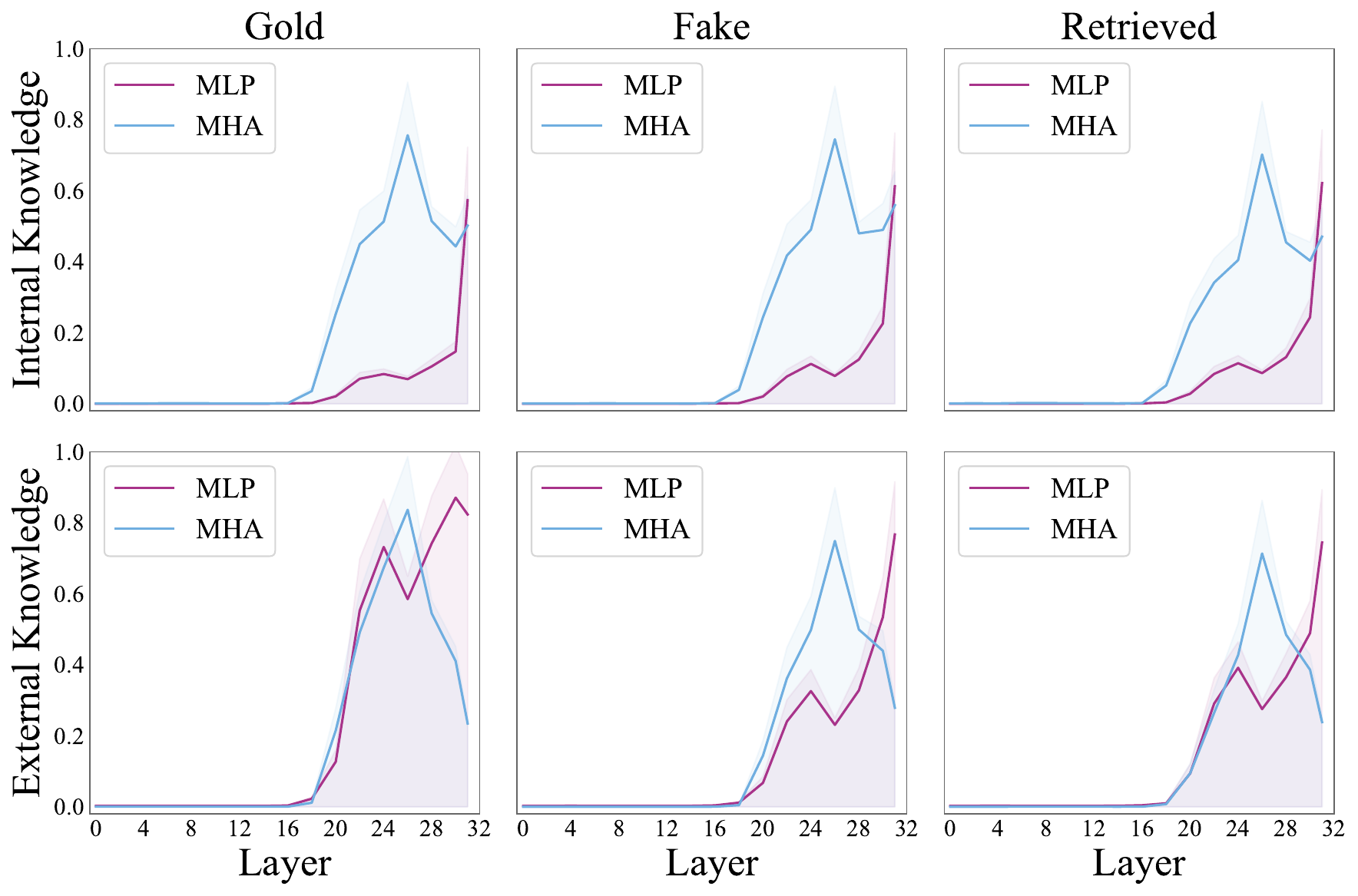}
	\caption{Unembedded logits of internal and external knowledge decoded at each layer with gold or fake passages as the external knowledge sources. }
    \vspace{-2ex}

	\label{fig:unembed}
\end{figure}

\subsubsection{Experiments and Analysis}
\label{sec:unembed-res}
We assess the unembedding logits of each layer’s MLP and MHA modules within the LLM to generate internal or external knowledge under RAG settings.
Figure~\ref{fig:unembed} shows the unembedding logits of internal or external knowledge using gold passages, fake passages and randomly selected retrieved passages~(explained in Section~\ref{sec:settings}) as external knowledge sources. The results are shown in Figure~\ref{fig:unembed}.

\textbf{First, we observe the contestation between internal and external knowledge in the deeper layers.} Our experiment reveals that both types of knowledge begin to manifest in the middle layers of the residual stream. This indicates that the parameters of the MLP and MHA modules start to contribute significantly to the formation of internal and external knowledge at this stage. This corroborates our delineation of the knowledge expression and contestation stages as described in Section~\ref{sec:phase} of the RAG LLMs.

\textbf{Moreover, we observe that the MLP is particularly sensitive to the correctness of external knowledge}. By comparing the unembedding logits between the results of gold passages and fake passages, we find a notable decline in the MLP’s contribution to the formation of external knowledge when the source is switched from the gold passage to the fake passage. Given that the only difference between the gold and fake passages lies in the core short answer, this suggests that the MLP is highly sensitive to the fine-grained accuracy of external knowledge. In contrast, no significant change in logits is observed in the MHA layers under similar conditions.

\textbf{Furthermore, we validate the role of the MLP in selecting between different knowledge sources, corroborating the rationale behind our neuron deactivation strategy.} We observe that the logits of internal knowledge show negligible differences across various external documents. However, the contribution of the MLP layer varies significantly. By deactivating the neurons responsible for these variations, we can alter the model's tendency to utilize internal versus external knowledge. This provides further insight into designing controllable RAG systems.

%% file: sec/sec-conclusion.tex
\section{Conclusion}

In this study, we present a novel investigation of LLM's knowledge utilization on how to integrate internal (parametric) and external (non-parametric) knowledge within RAG frameworks.
We analyze from two perspectives: macroscopic knowledge streaming and microscopic module functions.
At the macroscopic level, we identify four distinct stages that show how internal and external knowledge are generated and interact with each other. Our empirical findings demonstrate that the relevance of retrieved evidence is central to steering the knowledge elicitation stage.
At the microscopic level, we introduce the method of knowledge activation probability entropy (KAPE), to identify knowledge-specific neurons.
Leveraging this metric, we show that selectively deactivating these neurons effectively modulates the LLM’s reliance on each knowledge type.
Further analyses highlight the complementary roles of the multi-head attention (MHA) and multi-layer perceptron (MLP) modules. The MHA module facilitates the integration of information from multiple sources. Meanwhile, the MLP layers help ensure factual consistency in the final output.
These insights not only advance the interpretability of RAG systems, but also pave the way for designing more efficient and controllable LLM architectures capable of balancing parametric and contextual knowledge. 

Future work will focus on refining these mechanisms to improve real-world applications and extend the boundaries of AI-driven knowledge retrieval systems, developing more nuanced models that better handle the complexities of knowledge integration.